\documentclass[letterpaper, 10 pt, conference]{ieeeconf}
\IEEEoverridecommandlockouts    %
\usepackage{graphics}           
\usepackage{times}              
\usepackage{amsmath}            
\usepackage{amssymb}            
\usepackage{graphicx}
\usepackage{algorithm}
\usepackage{multirow}
\usepackage{url}
\usepackage[noend]{algpseudocode}
\usepackage{booktabs}
\usepackage{color}
\definecolor{instructioncolor}{rgb}{.5,.5,.5}

\usepackage[font=small]{caption}

\def\figref#1{Fig.~\ref{#1}}
\def\tabref#1{Tab.~\ref{#1}}
\def\eqref#1{Eq.~(\ref{#1})}

\makeatletter
\usepackage{xspace}
\DeclareRobustCommand\onedot{\futurelet\@let@token\@onedot}
\def\@onedot{\ifx\@let@token.\else.\null\fi\xspace}

\def\etal{{et al}\onedot}
\makeatother

\def\etalcite#1{\etal~\cite{#1}}

\usepackage{array}
\newcolumntype{L}[1]{>{\raggedright\let\newline\\\arraybackslash\hspace{0pt}}m{#1}}
\newcolumntype{C}[1]{>{\centering\let\newline\\\arraybackslash\hspace{0pt}}m{#1}}
\newcolumntype{R}[1]{>{\raggedleft\let\newline\\\arraybackslash\hspace{0pt}}m{#1}}

\newcommand{\RR}{\mathbb{R}}

\title{\LARGE \bf Radar Velocity Transformer:\\ Single-scan Moving Object Segmentation in Noisy Radar Point Clouds}

\author{Matthias Zeller \and Vardeep S. Sandhu \and Benedikt Mersch \and Jens Behley \and Michael Heidingsfeld \and Cyrill Stachniss%
  \thanks{Matthias Zeller and Vardeep S. Sandhu are with CARIAD SE and with the University of Bonn, Germany. Jens Behley and Benedikt Mersch are with the University of Bonn, Germany. Michael Heidingsfeld is with CARIAD SE, Germany. Cyrill Stachniss is with the University of Bonn, with the Department of Engineering Science at the University of Oxford, UK, and with the Lamarr Institute for Machine Learning and Artificial Intelligence, Germany.}%
}

\begin{document}
\maketitle
\thispagestyle{empty}
\pagestyle{empty}

\begin{abstract}
The awareness about moving objects in the surroundings of a self-driving vehicle is essential for safe and reliable autonomous navigation. The interpretation of LiDAR and camera data achieves exceptional results but typically requires to accumulate and process temporal sequences of data in order to extract motion information. In contrast, radar sensors, which are already installed in most recent vehicles, can overcome this limitation as they directly provide the Doppler velocity of the detections and, hence incorporate instantaneous motion information within a single measurement. %
In this paper, we tackle the problem of moving object segmentation in noisy radar point clouds. We also consider differentiating parked from moving cars, to enhance scene understanding.
Instead of exploiting temporal dependencies to identify moving objects, we develop a novel transformer-based approach to perform single-scan moving object segmentation in sparse radar scans accurately. 
The key to our Radar Velocity Transformer is to incorporate the valuable velocity information throughout each module of the network, thereby enabling the precise segmentation of moving and non-moving objects. Additionally, we propose a transformer-based upsampling, which enhances the performance by adaptively combining information and overcoming the limitation of interpolation of sparse point clouds.
Finally, we create a new radar moving object segmentation benchmark based on the RadarScenes dataset and compare our approach to other state-of-the-art methods. Our network runs faster than the frame rate of the sensor and shows superior segmentation results using only single-scan radar data. 
\end{abstract}

\section{Introduction}
\label{sec:intro}

\begin{figure}[t]
  \centering
  \fontsize{10pt}{10pt}\selectfont
     \def\svgwidth{\linewidth}
     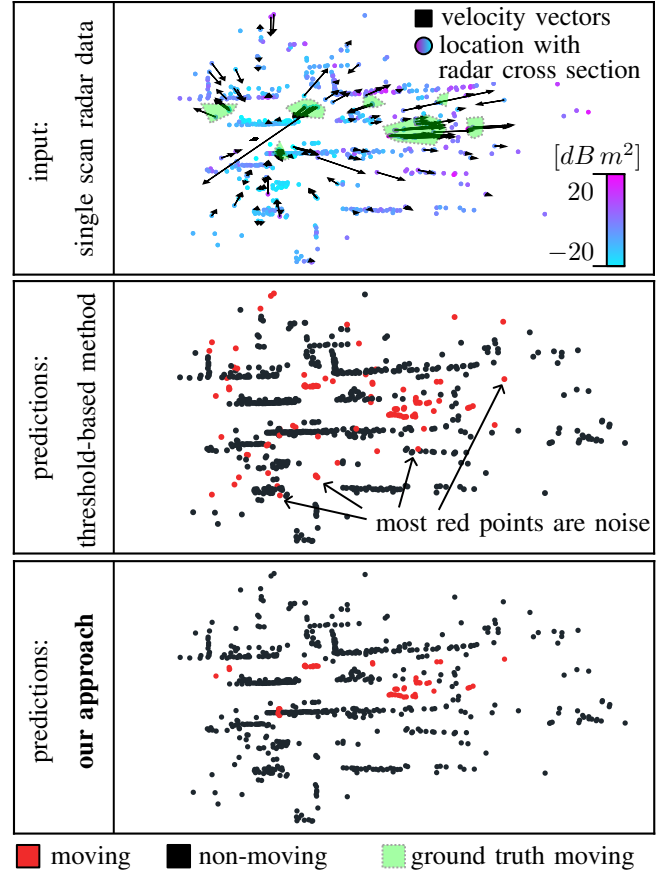
  \caption{Our learning-based approach enhance moving object segmentation (middle), from noisy, single-scan radar point clouds (top) compared to a velocity threshold determined on the validation set. Best viewed in color.}
  \vspace{-0.2cm}
  \label{fig:motivation}
  \vspace{-0.7cm}
\end{figure}

Self-driving vehicles need to distinguish moving from stationary objects to safely navigate in dynamic, real-world environments. To enable redundancy and overcome the shortcomings of individual sensors, the sensor suites of autonomous vehicles are versatile, including cameras, LiDARs, and radars. The widely explored camera and LiDAR sensors utilize temporal sequences of input data to segment moving objects, often neglecting the valuable information of radar data. The Doppler velocity provided by a radar enables the identification of moving objects in single scans and radar sensors work under adverse weather, including rain, fog, and snow where other modalities encounter difficulties. A serious drawback, however, is that the radar scans are largely affected by noise due to multi-path propagation, ego-motion, and sensor noise. The noisy measurements frequently lead to false positives and make threshold-based moving object segmentation~\cite{scheiner2021aip} unacceptable, as visualized in~\figref{fig:motivation}. We aim at investigating in this paper how the additional sensor information of the Doppler velocity can be exploited by learning-based approaches to enable reliable identification of moving objects in the environment. Furthermore, the radar cross section, which depends on the material properties and the structure of the detection, supports the differentiation of closely connected objects.

We investigate moving object segmentation in radar point clouds. This task requires differentiation between the detection of moving and stationary objects. To accurately differentiate between the two classes, we exploit single radar point clouds, including the valuable Doppler velocity and radar cross section.
State-of-the-art methods for moving object segmentation for camera and LiDAR data rely on the elaboration of temporal dependencies in videos~\cite{fan2021cvpr} or aggregated residuals between previous scans~\cite{kim2022ral}. The processing of multiple frames induces latency which is unsuitable for a task requiring immediate information about the environment such as collision avoidance. Therefore, we investigate the processing of single, sparse radar point clouds by exploiting additional and valuable radar sensor information to leverage the full potential of radar sensors.

The main contribution of this paper is a novel learning-based approach that accurately predicts moving objects in sparse, single-scan radar point clouds. Our approach, called Radar Velocity Transformer, predicts for each point in the input radar scan the semantic label of moving or non-moving. To classify the individual detection and extract valuable point-wise features, we introduce the velocity encoding in each module of our network. The encoding of the velocity enhances the performance by injecting important information throughout the network. We optimize the feature aggregation in the decoder part by our transformer-based upsampling to adaptively merge features and capture complex local structures in sparse point clouds. Furthermore, we reorganized the RadarScenes~\cite{schumann2021icif} dataset providing semantic classes for individual detection, which we transfer into moving and non-moving labels establishing a single-scan benchmark.

In sum, we make two key claims:
(i) Our approach is able to accurately perform moving object segmentation in single-scan, noisy radar point clouds and enhance the state of the art in moving object segmentation without exploiting temporal dependencies;
(ii) The velocity encoding throughout the network and the transformer-based upsampling are essential to derive highly discriminative features and adaptively aggregate information to enhance accuracy. 

\begin{figure*}[t]
 \centering
 \fontsize{8pt}{8pt}\selectfont
 \def\svgwidth{\textwidth}
 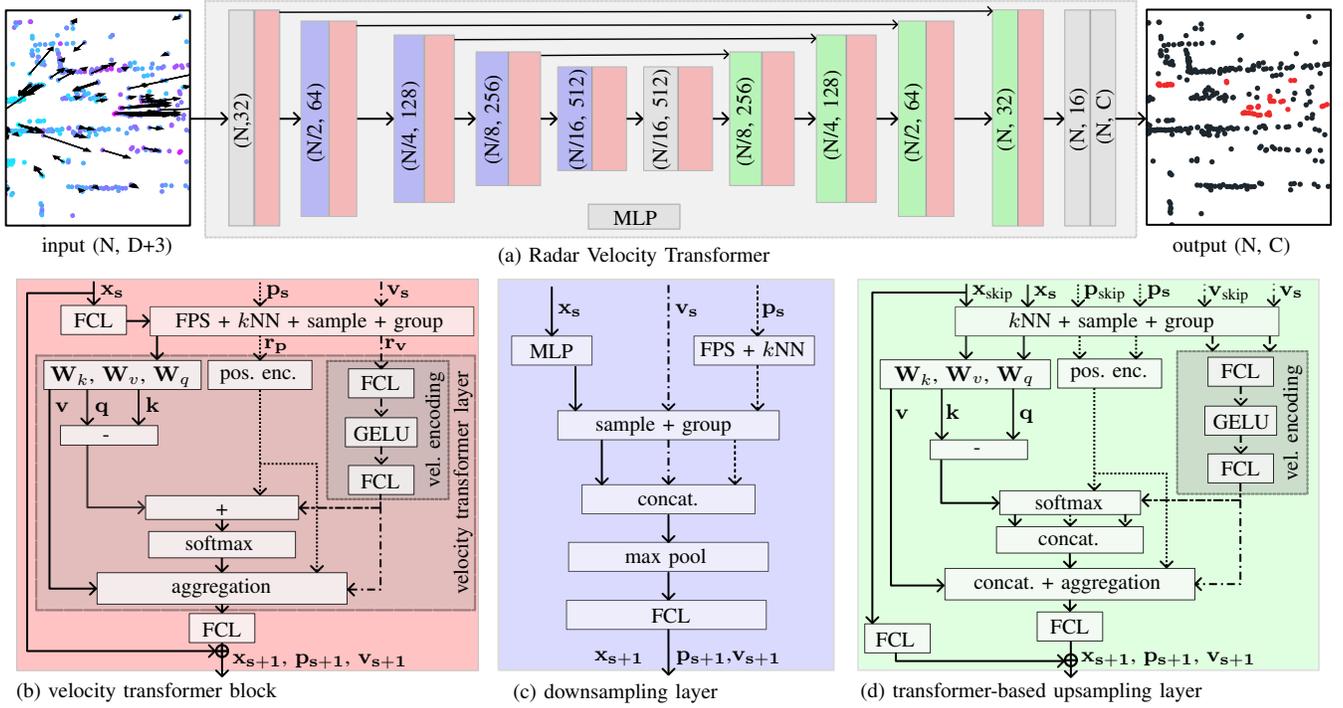
 \caption{The detailed design of each module of our Radar Velocity Transformer (a) shows the Radar Velocity Transformer, (b) the velocity transformer block with the velocity transformer layer, (c) the downsampling layer, and (d) the transformer-based upsampling layer. The different colors stand for the different building blocks. The tuples denote the number of points and feature channels in each stage. MLP: multi-layer perceptron, FCL: fully connected layer, vel. encoding: velocity encoding, pos. enc.: positional encoding, FPS: farthest point sampling, $k$NN: $k$-nearest neighbor, concat.: concatenation, $W$: weight matrices}
 \vspace{-0.1cm}
 \label{fig:modules}
  \vspace{-0.5cm}
\end{figure*}

\section{Related Work}
\label{sec:related}

Moving object segmentation in point clouds can be categorized into map-based~\cite{kim2020iros,pagad2020icra,ruchti2018icra} and map-free approaches \cite{kim2022ral,mersch2022ral,sun2022mos3d}. Current advancements focus on the latter to work online and remove the burden of pre-built maps, which is further supported by scene flow estimation~\cite{baur2021iccvn,ding2022ral,gu2019cvprn,kittenplon2021cvpr,shi2022icra,wei2021cvpr,wu2020eccv} and semantic segmentation~\cite{guo2021pct,kaul2020iv,lai2022cvpr,xu2021iccv,xu2021cvpr,zhao2021iccv}. To differentiate between the methods, we distinguish between projection-based, voxel-based, point-based, transformer-based, and hybrid methods.

\textbf{Projection-based} methods are introduced to utilize the successful convolutional neural networks~(CNNs) on 3D data. For example, Chen~\etalcite{chen2021ral} first project the LiDAR point clouds into 2D range images and provide the residual images of previous scans as input to SalsaNext~\cite{cortinhal2020svc} to perform moving object segmentation on SemanticKITTI~\cite{behley2019iccv}. Kim~\etalcite{kim2022ral} extend the approach and improve state-of-the-art performance for moving object segmentation in LiDAR data by efficient data augmentation and the attention-based fusion module to combine semantic and motion features. The methods are highly efficient but face back projection artifacts when transferring the 2D predictions to the 3D point cloud which harms the accuracy. 

\textbf{Voxel-based} methods keep the 3D information intact and hence reduce the limitations caused by back projection. Mersch~\etalcite{mersch2022ral} adopted the Minkowski engine~\cite{choy2019cvpr} and propose a receding horizon strategy to incorporate new scans in an online fashion and refine predictions by Bayesian filtering to enhance LiDAR moving object segmentation.
However, the voxel-based methods inherently introduced discretization artifacts resulting in an information loss.

\textbf{Point-based} methods~\cite{liu2019cvpr,qi2017cvpr,wang2020wacv} allow for keeping the full spatial information of point clouds, which is desirable, especially for sparse point clouds. The pioneering work of Qi~\etalcite{qi2017cvpr}, called PointNet, utilizes shared multi-layer perceptrons~(MLPs) to directly consume point clouds and aggregate nearby information by max pooling functions. FlowNet3D~\cite{liu2019cvpr}, FlowNet3D++~\cite{wang2020wacv}, and FLOT~\cite{puy2020eccv} follow PointNet++~\cite{qi2017nips2} and introduce dedicated architectures for point-based scene flow estimation in LiDAR point clouds. Schumann~\etalcite{schumann2018icif} adapt the hierarchical grouping of PointNet++~\cite{qi2017nips2} combining features from multiple scales and extending their approach by exploiting strong temporal dependencies by the aggregation of consecutive scans~\cite{schumann2020tiv} to enhance semantic segmentation of sparse radar point clouds. Fan~\etalcite{fan2021cvpr} propose P4Transformer, which combines 4D point-based convolutions with video-level self-attention to merge related local areas spatially and temporally. The point-based method benefits from the transformer-based module since the self-attention mechanism ~\cite{paigwar2019cvprws,xie2018cvpr,yang2019cvpr} is invariant to permutation and thus inherently suitable to capture strong local and global dependencies and extract valuable features in point clouds. 

\textbf{Transformer-based} methods dominate a variety of tasks from natural language processing to point cloud understanding by exploiting the powerful self-attention mechanism~\cite{dai2019acl,guo2021pct,kolesnikov2021iclr,ramachandran2019nips,vaswani2017nips,lai2022cvpr,zeller2023ral,zhao2021iccv}. Guo~\etalcite{guo2021pct} propose offset-attention with an implicit Laplace operator and normalization refinement to reduce the influence of noise and sharpen the attention weights. SAFIT~\cite{shi2022icra} models relations on object and point level via transformers to estimate scene flow. Since the self-attention mechanism is computationally expensive, transformer-based networks benefit from effective sampling strategies to aggregate local features and reduce the computational cost. Besides the wide range of sampling algorithms~\cite{hu2020cvpr,wu2019cvpr,yang2020ijcv,yang2019cvpr} the common method for downsampling is farthest point sampling following max pooling~\cite{qi2017nips2}. For upsampling, trilinear interpolation is usually the method of choice based on an inverse distance weighted average~\cite{qi2017nips2}. 
To further keep fined-grained position information throughout the network, Zhao~\etalcite{zhao2021iccv} introduce the trainable position encoding and adapt vector-based attention~\cite{zhao2020cvpr}. Stratified Transformer~\cite{lai2022cvpr} extends the position encoding and aggregates long-range context by a window-based key-sampling strategy to enhance the accuracy.

In this paper, we follow recent advancements and propose a novel transformer-based moving object segmentation method for sparse and noisy radar data. In contrast to the related work, our newly introduced Radar Velocity Transformer extends the transformer layer and exploits the valuable velocity information throughout the network. Furthermore, our proposed network includes an advanced transformer-based upsampling strategy to capture complex local structures and enhances state-of-the-art performance for moving object segmentation in single-scan radar point clouds. 
\section{Our Approach}
\label{sec:main}
The goal of our approach is to achieve precise moving object segmentation in single-scan, sparse radar point clouds to enhance the environmental perception of autonomous vehicles. \figref{fig:modules} illustrates our Radar Velocity Transformer~(RVT), which is a transformer-based framework that builds upon the successful self-attention mechanism~\cite{vaswani2017nips} and directly processes the input point cloud to omit information loss. We incorporate the valuable Doppler velocity information within each module and use the so-called velocity transformer layer as the central building block of each encoder-decoder stage. Furthermore, we introduce transformer-based upsampling modules to adaptively combine local context information to enable fine-grained feature extraction.

\subsection{Velocity Transformer Layer}
\label{sec:rvtl}

In sparse radar point clouds, the information of individual detections can be of great benefit for solving downstream tasks such as moving object segmentation. Therefore, we introduce a velocity transformer layer to enhance feature extraction, as illustrated in \figref{fig:modules}~(b). 

The inputs are single-scan radar point clouds $\mathcal{P}_{s}$ with point coordinates $\mathbf{p}_i \in \RR^{2}$, ego-motion compensated Doppler velocity $\mathbf{v}_i \in \RR$, and point-wise features $\mathbf{x}_i\in\RR^{D}$ with feature dimension,~$D$.

Since the Doppler velocity provides essential information about the moving and non-moving parts of the environment, we incorporate the information as a central part of our velocity transformer layer. The idea is to support accurate moving object segmentation based on the relative velocity $\mathbf{r}^{v}_{i,j}\in\RR^{N\times N_{\text{vtl}}}$ since this enables the differentiation of nearby points and the identification of moving objects. Hence, we process the relative velocity $\mathbf{r}^{v}_{i,j}=\mathbf{v}_i-\mathbf{v}_j$ by two fully connected layers and the Gaussian error linear unit~(GELU) as an activation function~\cite{hendrycks2016corr} to include the information.
The rest of our velocity transformer layer follows standard transformers~\cite{zhao2020cvpr,zhao2021iccv} and relies on the encoded representation of the input features $\mathbf{x}$. The queries $\mathbf{q}$, the keys $\mathbf{k}$, and the values $\mathbf{v}$ are determined by multi-layer perceptrons with the corresponding weight matrices $\mathbf{W}_{q} \in \RR^{D\times D}$, $\mathbf{W}_{k} \in \RR^{D\times D}$ and $\mathbf{W}_{v} \in \RR^{D\times D}$, as follows:
\begin{align}
\mathbf{q} &= \mathbf{W}_q \mathbf{x}, &
\mathbf{k} &= \mathbf{W}_k \mathbf{x}, &
\mathbf{v} &= \mathbf{W}_v \mathbf{x}.
\label{eq:1}
\end{align}

As relation functions $\mathbf{g}$ for the queries and the keys, we utilize subtraction.
For the positional encoding, we adapt the approach of Zhao~\etalcite{zhao2021iccv} and process the relative position $\mathbf{r}^{p}_{i,j}=\mathbf{p}_i-\mathbf{p}_j$ by two fully connected layers and the GELU. To calculate the attention scores $\mathbf{a}_{i,j}$ within local areas, we adapt vector attention~\cite{zhao2020cvpr} to allow for a weighting of individual feature channels. We determine the local areas with $N_{\text{vtl}}$ points by farthest point sampling and $k$-nearest neighbor~($k$NN) algorithm.
To enable fine-grained information aggregation, we calculate attention weights based on the sum of the relation of queries and keys $\mathbf{g}(\mathbf{q}_{i},\mathbf{k}_{j})$, the relative position encoding $\mathbf{r}^{p}_{i,j}$, and the relative velocity encoding $\mathbf{r}^{v}_{i,j}$.
The final attention weights are determined by the softmax function:
\begin{align}
\mathbf{a}^{i,j}= \text{softmax}(\mathbf{g}(\mathbf{q}_{i},\mathbf{k}_{j})+\mathbf{r}^{p}_{i,j}+\mathbf{r}^{v}_{i,j}).
\label{eq:2}
\end{align}

Additionally, we add the relative velocity encoding to the values and the relative position encoding to derive the combined values $\mathbf{v}^{c}_{i,j}=\mathbf{v}_{i,j}+\mathbf{r}^{p}_{i,j}+\mathbf{r}^{v}_{i,j}$, which include and update the valuable information throughout the network. To derive the weighted features $\mathbf{y}$, we calculate the sum of the element-wise multiplication:
\begin{align}
\mathbf{y}_j &= \sum_{i=1}^{N_{\text{vtl}}}{ \mathbf{a}_{i,j}\odot \mathbf{v}^{c}_{i} },
\end{align}
within the local areas. The aggregated features which are enriched by the velocity encoding $\mathbf{y}$ are directly processed by the following module to reduce the computational cost within the velocity transformer layer.

\subsection{Velocity Transformer Block}
\label{sec:rvtb}
Our velocity transformer block is a residual block~\cite{he2016cvpr}, similar to the point transformer block~\cite{zhao2021iccv},  that embeds the velocity transformer layer in the center of two fully connected layers, as depicted in \figref{fig:modules}~(b). We add LayerNorm~\cite{xiong2020icml} and a GELU activation function for each fully connected layer. The features $\mathbf{x}_i$ are processed by the velocity transformer layer and the linear layers to enrich the information of individual points within local areas. The velocity and position data are utilized to determine the relative encodings but are not further transformed to keep unaltered information throughout the network.

\subsection{Downsampling Layer}
\label{sec:ds}
The downsampling layer reduces the cardinality of the point cloud $\mathcal{P}_{s+1} \subset \mathcal{P}_{s}$ after each stage $s$ and has to keep the most relevant information intact. Following Qi~\etalcite{qi2017nips2}, we adapt the max pooling operation depicted in \figref{fig:modules}~(c). We first process the feature vector by a linear layer. To derive the local areas, we sample and group the points by farthest point sampling and $k$NN algorithm. The features and the Doppler velocity values are sampled and grouped accordingly. To also induce valuable velocity information in the downsampling, we concatenate the features, the position, and the velocity information. Afterward, we apply max pooling to aggregate the information and process the feature vector by a fully connected layer. We reduce the number of points $N_s$ by a factor of 2 and keep the position and velocity information of the downsampled point cloud to enrich the information in deeper layers. 

\subsection{Transformer-based Upsampling Layer}
\label{sec:us}
The common upsampling method interpolates the $k=3$ nearest neighbors based on an inverse distance weighted average~\cite{qi2017nips2} and combines these with the features of the skip connection. Especially at the boundaries of moving objects, the straightforward interpolation can result in a combination of features of different classes, which can harm the extraction of discriminative features. Hence, we argue that the upsampling and the aggregation of the features in the decoder part of the network are crucial to enhance accuracy, especially for sparse point clouds. 

To adaptively merge the information of the two point clouds, we propose the transformer-based upsampling layer visualized in \figref{fig:modules}~(d). The idea is to enable the network to learn to concatenate important information by inter-attention to extract valuable features. The inputs are the output point cloud of the previous velocity transformer block $\mathcal{P}_{s}$, with the number of points $N_s$, which has to be upsampled, and the point cloud of the skip connection $\mathcal{P}_{\text{skip}}$ where $N_{s} \leq N_{\text{skip}}$. 

Inspired by our velocity transformer layer, we first encode the features $\mathbf{x}_{s}$ as keys $\mathbf{k}$, and values $\mathbf{v}$ and the features $\mathbf{x}_{\text{skip}}$ as queries $\mathbf{q}$, following \eqref{eq:1}. To determine the relative position and velocity encoding, we calculate the $k$-nearest neighbors for the point cloud of the skip connection $\mathcal{P}_{\text{skip}}$ within the point cloud $\mathcal{P}_{s}$. 

In the sample and grouping module, we compute the relative position and velocity of the correspondent points of the two point clouds. We determine the encodings by two fully connected layers with the GELU activation function. In contrast to the velocity transformer layer, we calculate individual attention weights for the relation of queries and keys $\mathbf{a}^{qk}_{i,j}$, the relative position encoding $\mathbf{a}^{p}_{i,j}$, and the relative velocity encoding $\mathbf{a}^{v}_{i,j}$ to enable fine-grained information aggregation and enhance accuracy.
The individual attention weights are determined by the softmax function as follows:
\begin{align}
\mathbf{a}^{qk}_{i,j}=& \text{softmax}(\mathbf{g}(\mathbf{q}_{i},\mathbf{k}_{j})),\\
\mathbf{a}^{p}_{i,j}=& \text{softmax}(\mathbf{r}^{p}_{i,j}),\\
\mathbf{a}^{v}_{i,j}=& \text{softmax}(\mathbf{r}^{v}_{i,j}).
\label{eq:4}
\end{align}

We concatenate the individual attention weights to derive the final attention scores $\mathbf{a}_{i,j}=(\mathbf{a}^{qk}_{i,j},\mathbf{a}^{p}_{i,j},\mathbf{a}^{v}_{i,j})$. To weight the respective information, the values $\mathbf{v}$ are concatenated with $\mathbf{r}^{p}_{i,j}$ and the velocity encoding $\mathbf{r}^{v}_{i,j}$ resulting in the combined values $\mathbf{v}^{c}_{i,j}=(\mathbf{v}_{i,j},\mathbf{r}^{p}_{i,j},\mathbf{r}^{v}_{i,j})$. 

To derive the weighted features $\mathbf{y}$, we calculate the sum of the element-wise multiplication:
\begin{align}
\mathbf{y}_j &= \sum_{i=1}^{N_{\text{tus}}}{ \mathbf{a}_{i,j}\odot \mathbf{v}^{c}_{i} },
\end{align}
within local areas. 
The aggregated features $\mathbf{y}$ are processed by a fully connected layer to compress the features to the original feature dimension~$D$ with a learnable weight matrix $\mathbf{W}_y \in \RR^{(D+12)\times D}$: 
\begin{align}
\mathbf{z} &= \mathbf{W}_y \mathbf{y}, 
\label{eq:5}
\end{align}
where $\mathbf{z}$ are the updated features for the upsampled point cloud. The fully connected layer enables the information exchange of the individual parts and reduces the complexity of the succeeding modules.

The final output of the transformer-based upsampling layer is the sum of the features $\mathbf{x}_{\text{skip}}$ and $\mathbf{z}$, which incorporates the valuable information of both point clouds to derive discriminative features.

\subsection{Network Architecture}
\label{sec:backbone}
We build our network architecture based on the widely-used U-Net~\cite{qi2017nips2,zhao2021iccv} with an encoder-decoder architecture including skip connections as illustrated in \figref{fig:modules}. The input to the network are the features $\mathbf{x}^i_i$, the position information~$\mathbf{p}_{i}$ with two spatial coordinates $x^C_{i}$, $y^C_{i}$, and the ego-motion compensated Doppler velocity $v_i$. The features include the position, the velocity, and additionally, the radar cross section~$\sigma_i$ resulting in a 4-dimensional vector $\mathbf{x}^i_i=(x^C_i, y^C_i, v_i, \sigma_i)$. The input $\mathbf{x}^i_i$ are processed in each stage $s$ resulting in the features $\mathbf{x}_s$. The input is first processed by an MLP before being passed to the first velocity transformer layer. The per-point features are gradually increased within each stage from 32 to 64, 128, 256, and 512. The sampling operations change the cardinality by a factor of 2 resulting in [N, N/2, N/4, N/8, N/16] points for the respective stage. The final output is determined by an MLP with two linear layers to obtain per-point logit values. The individual stages of our architecture each comprise one single velocity transformer block to build an efficient network. 
\subsection{Implementation Details}
\label{sec:impl}
The Radar Velocity Transformer is implemented in PyTorch~\cite{paszke2019nips}. We train our model over 50 epochs with AdamW~\cite{loshchilov2017iclr} optimizer with an initial learning rate of 0.0005 and a cosine annealing learning rate scheduler~\cite{loshchilov2017iclr}. We combine the Lovász loss~\cite{berman2018cvpr} and the weighted cross-entropy for which we follow the approach by Schumann~\etalcite{schumann2020tiv} and set the weights for moving objects to 8.0 and for static ones to 0.5. The local areas for the velocity transformer layer are set to $N_{\text{vtl}}=16$ and for the transformer-based upsampling to $N_{\text{tus}}=12$. 
We train the network with one Nvidia A100 GPU and a batch size of 128. To reduce overfitting, we further apply data augmentation, using jitter, scaling, rotation, and instance augmentation.

\begin{table}[t]
  \centering
  \begin{tabular}{@{}llc@{}}
\toprule
Method                     & Input         & IoU           \\ \midrule
Threshold $\lvert v_i\rvert >t$                  & single-scan   & 35.1          \\
4DMOS~\cite{mersch2022ral}                      & multiple-scan & 73.1          \\
Stratified Transformer~\cite{lai2022cvpr}     & single-scan   & 74.6          \\
Our Radar Velocity Transformer & single-scan   & \textbf{81.3} \\ \bottomrule
\end{tabular}%
  \caption{Moving object segmentation results on the RadarScenes test set in terms of IoU for the moving class. The threshold \mbox{$t=0.92\,m/s$} is determined on the validation set and afterwards applied to the test set~\cite{scheiner2021aip}.\vspace{-0.2cm}}
  \label{tab:perf}
  \vspace{-0.2cm}
\end{table}

\section{Experimental Evaluation}
\label{sec:exp}

The main focus of this work is an accurate, single-scan moving object segmentation in sparse and noisy radar point clouds.
We present our experiments to show the capabilities of our method to reliably segment moving objects. The results of our experiments also support our key claims, which are: Our approach (i)~segments moving objects in radar point clouds more precisely compared to state-of-the-art methods and (ii)~the velocity encoding and the transformer-based upsampling enhance the accuracy by incorporating valuable information throughout the network and.

\subsection{Experimental Setup}
We utilize the RadarScenes~\cite{schumann2021icif} dataset, to train and evaluate our model since this dataset is the only open, large-scale radar dataset that includes per-point annotations for moving objects under different weather conditions and driving scenarios. The dataset is split into 130 sequences for training and 28 for validation. To construct a test set, we split the RadarScenes validation set into 6 sequences for validation (6, 42, 58, 85, 99, 122) and the remaining 22 sequences for testing. 

The RadarScenes~\cite{schumann2021icif} dataset includes four radar sensors. To provide information on the surroundings of the vehicle, we need to merge the point clouds of the individual sensors into one central radar scan. Since the pose information, the measurement time, and the coordinates of the individual detection are given, we merge the data within a common coordinate system.

Following Chen~\etalcite{chen2021ral}, we utilize the intersection over union~(IoU)~\cite{everingham2010ijcv}, where $IoU=\frac{TP}{TP+FN+FP}$ with the number of true positive~(TP), false positive~(FP), and false negative~(FN) predictions for moving objects to evaluate the methods.

\begin{figure*}[t]
  \centering
  \fontsize{8pt}{8pt}\selectfont
     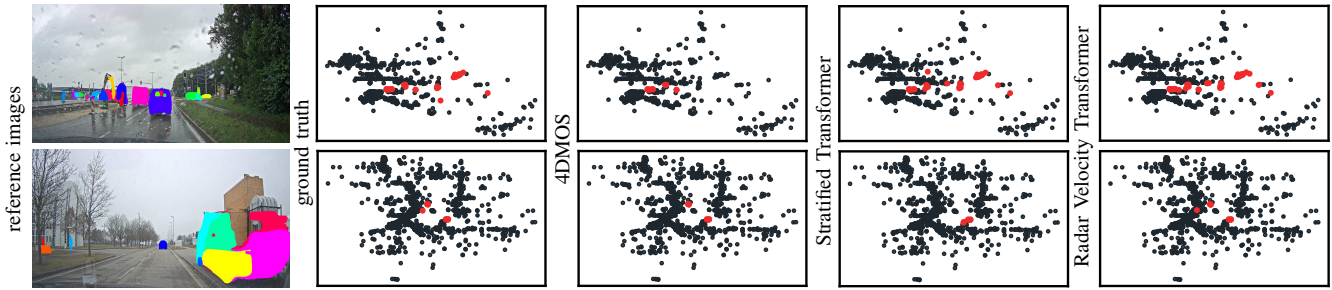
  \caption{Qualitative results of 4DMOS~\cite{mersch2022ral}, Stratified Transformer~\cite{lai2022cvpr}, and our Radar Velocity Transformer on the test set of RadarScenes~\cite{schumann2021icif}. Red points indicate moving objects and black points belong to static objects.}
    \vspace{-0.1cm}
  \label{fig:resw}
  \vspace{-0.2cm}
\end{figure*}
\subsection{Moving Object Segmentation Performance}
\label{sec:perf}
The first experiment evaluates the performance of our approach and its outcome supports the claim that our approach enhances state-of-the-art moving object detection in sparse and noisy radar point clouds utilizing only single scans.

To compare the results, we select the recently best-performing point-based segmentation method, the Stratified Transformer~\cite{lai2022cvpr}, which utilizes single scans, the 4DMOS network~\cite{mersch2022ral} for LiDAR moving object segmentation which does not use the range representation because this is incompatible with the 2D coordinates, and a simple threshold for the velocity determined on the validation set~\cite{scheiner2021aip}. For specific information on the training regime of the two networks, we refer to the original papers~\cite{mersch2022ral,lai2022cvpr}.

The Radar Velocity Transformer outperforms the two existing approaches and the learning-based methods are superior compared to the threshold-based method, as displayed in \tabref{tab:perf}. The difference between the learning-based approaches and the threshold-based method illustrates the necessity of advanced models to perform moving object segmentation in noisy radar point clouds. Additionally, the transformer-based methods enhance the performance compared to the voxel-based 4DMOS, which suggests that discretization artifacts lead to information loss that cannot be compensated by additional temporal information of consecutive radar scans. The feature input vector of Stratified Transformer and Radar Velocity Transformer both contain valuable velocity information. However, our Radar Velocity Transformer considerably improves the IoU for moving objects by 6.7 absolute percentage points and performs well under adverse weather conditions, as illustrated in~\figref{fig:resw}.

\begin{table}[t]
  \centering
  \begin{tabular}{@{}llll@{}}
\toprule
\#      & \begin{tabular}[c]{@{}l@{}}velocity \\ encoding\end{tabular} & \begin{tabular}[c]{@{}l@{}}transformer-based\\ upsampling\end{tabular} & IoU  \\ \midrule
{[}A{]} &                                                              &                                                                        & 73.4 \\
{[}B{]} &                                                              &    \checkmark                                                                    & 75.2 \\
{[}C{]} &            \checkmark                                                  &                                                                        & 75.6 \\
{[}D{]} &         \checkmark                                                     & \checkmark                                                                       & \textbf{77.4} \\ \bottomrule
\end{tabular}%

  \caption{Influence of the different components in terms of IoU for moving objects on the RadarScenes validation set.}
  \label{tab:components}
  \vspace{-0.2cm}
\end{table}
\subsection{Ablation Study on Network Components}
The second experiment, the ablation study on network components, evaluates the influence of the velocity encoding and transformer-based upsampling on the performance to support our second claim that our proposed modules each contribute to the improvements in terms of IoU. The combined results of the ablation study on the validation set are listed in~\tabref{tab:components}.

To assess the benefits of transformer-based upsampling, we replace the module with the commonly used trilinear interpolation based on an inverse distance weighted average~\cite{qi2017nips2}. Since the velocity encoding is new and the information of the velocity of the individual detection is present in the feature vector $\mathbf{x}$, we remove the velocity encoding to evaluate the influence on the IoU.

Ablation [A], we replace the upsampling and remove the velocity encoding which leads to a decrease in terms of IoU by 4 absolute percentage points. In ablation [B], we add the transformer-based upsampling, which enables an adaptive feature aggregation of the two point clouds and leads to an improvement of IoU by 1.8 absolute percentage points. In comparison to ablation [A], we add the velocity encoding throughout the network in [C], which enhances the performance. We assume that the velocity encoding is highly valuable since the fine-grained Doppler velocity information may be lost in high-level features of deeper layers. Hence, the specific task of moving object segmentation benefits from the velocity encoding. The final model of our Radar Velocity Transformer, represented in [D], further enhances the IoU by the usage of both, velocity encoding and transformer-based upsampling. We conclude that the additional information of the velocity encoding supports the aggregation of the features for the upsampling and hence leads to the best results.

As an additional experiment, we replaced the concatenation of the transformer-based upsampling with an addition in our final Radar Velocity Transformer. The obtained IoU of 75.3\,\% indicates that the concatenation leads to a more fine-grained weighting of the individual channels and improves the performance. Additionally, we exploit transformer-based downsampling. However, this does not improve the overall performance and hence we keep the max pooling since it is more efficient and does not mix information, which is suitable for the downsampling of sparse point clouds.

\subsection{Runtime}
Finally, we analyze the runtime of our approach and show that our
approach runs fast enough to support online processing in the vehicle. We tested our approach on an AMD Ryzon~5 CPU with an Nvidia GTX 1660 GPU.
Our implementation includes an optimized farthest point sampling and $k$NN algorithm in C++ to speed up the inference. Since the point clouds differ in the number of detections, we evaluate 1,000 scans that are randomly selected from the validation set. The mean runtime is 0.012\,s, which is equal to 83\,Hz, and thus over 4x faster than the frame rate of 17\,Hz of the sensor.

\section{Conclusion}
\label{sec:conclusion}

In this paper, we presented a novel approach to accurately perform single-scan moving object segmentation in the domain of radar data. 
Our approach encodes the valuable Doppler velocity information throughout the network and optimizes the upsampling operation by adaptively aggregating information to enhance performance.
This allows us to successfully differentiate between moving and non-moving objects, which we evaluated on the RadarScenes dataset. The experiments and the comparisons to other approaches support all claims made in this paper and suggest that our architecture achieves superior performance on moving object segmentation in noisy, single-scan point clouds obtained from automotive radars.
The sensors used for recording the RadarScenes dataset are series sensors, already implemented in vehicles, which makes our approach available without additional cost.
Overall, our approach outperforms the state-of-the-art methods and proposes a new benchmark for radar-based moving object segmentation, which allows further comparisons with future work, taking a step forward towards reliable single-scan moving object segmentation and sensor redundancy for autonomous vehicles.

\bibliographystyle{plain_abbrv}

\bibliography{glorified,new}

\end{document}